\documentclass{article}
\usepackage{ijcai25}

\usepackage{times}
\usepackage{soul}
\usepackage{url}
\usepackage[hidelinks]{hyperref}
\usepackage[utf8]{inputenc}
\usepackage[small]{caption}
\usepackage{graphicx}
\usepackage{amsmath}
\usepackage{amsthm}
\usepackage{booktabs}
\usepackage{algorithm}
\usepackage{algorithmic}
\usepackage[switch]{lineno}
\usepackage{tabularray}
\title{NSF-MAP: \underline{N}euro\underline{s}ymbolic \underline{M}ultimodal \underline{F}usion for Robust and Interpretable \underline{A}nomaly \underline{P}rediction in Assembly Pipelines}

\author{
Chathurangi Shyalika$^{1*\dagger}$ \and
Renjith Prasad$^{1*}$ \and
Fadi El Kalach$^{2}$ \and
Revathy Venkataramanan$^{1}$ \and
Ramtin Zand$^{3}$ \and
Ramy Harik$^{2}$ \and
Amit Sheth$^{1}$ \\
\affiliations
$^1$Artificial Intelligence Institute, University of South Carolina \\
$^2$Clemson Composites Center, Clemson University \\
$^3$Intelligent Circuits, Architectures and Systems Lab, University of South Carolina \\
\emails
\{jayakodc, revathy, amit\}@email.sc.edu, kaippilr@mailbox.sc.edu, ramtin@cse.sc.edu, \\ \{felkala, harik\}@clemson.edu
}

\begin{document}

\maketitle
\let\thefootnote\relax
\footnotetext{$^*$Equal contribution. $^*$Corresponding author.}

\begin{abstract}
  In modern assembly pipelines, identifying anomalies is crucial in ensuring product quality and operational efficiency. Conventional single-modality methods fail to capture the intricate relationships required for precise anomaly prediction in complex predictive environments with abundant data and multiple modalities. This paper proposes a neurosymbolic AI and fusion-based approach for multimodal anomaly prediction in assembly pipelines. We introduce a time series and image-based fusion model that leverages decision-level fusion techniques. Our research builds upon three primary novel approaches in multimodal learning: time series and image-based decision-level fusion modeling, transfer learning for fusion, and knowledge-infused learning. We evaluate the novel method using our derived and publicly available multimodal dataset and conduct comprehensive ablation studies to assess the impact of our preprocessing techniques and fusion model compared to traditional baselines. The results demonstrate that a neurosymbolic AI-based fusion approach that uses transfer learning can effectively harness the complementary strengths of time series and image data, offering a robust and interpretable approach for anomaly prediction in assembly pipelines with enhanced performance. \noindent The datasets, codes to reproduce the results, supplementary materials, and demo\footnote{Video URL:\url{https://shorturl.at/IY9tt}} are available at \url{https://github.com/ChathurangiShyalika/NSF-MAP}.
\end{abstract}

\section{Introduction}

Anomaly prediction in industrial assembly pipelines is essential for maintaining product quality and operational efficiency. Traditional methods focusing on unimodal data \cite{chandola2009anomaly,kim2023self,shyalika2024ri2ap} often fall short in complex environments where anomalies are influenced by multiple variables. Multimodal anomaly prediction, which integrates various data modalities like time series, image, text, and video, provides a more comprehensive and accurate prediction mechanism by uncovering hidden patterns and relationships that single-modality methods often overlook \cite{Nedelkoski2019AnomalyDF,Zhao2020FailurePI}. This approach is particularly beneficial in manufacturing settings where variables like temperature, speed, and visual inspections are interrelated, necessitating advanced methods to handle multimodal and multivariate data effectively \cite{Zhao2020FailurePI,Nedelkoski2019AnomalyDF}. 
%Despite the benefits of multimodal anomaly prediction, challenges such as data scarcity, poor performance metrics, real-time integration challenges, and limited interpretability persist in manufacturing assembly processes, highlighting a need for more practical, scalable, and interpretable solutions \cite{mccormick2025real}.
Despite the benefits of multimodal anomaly prediction, challenges such as the extreme scarcity of datasets containing both time-series and image data—where we identified only one (details in Section \ref{sec:dp})—along with poor performance metrics, real-time integration difficulties, and limited interpretability persist in manufacturing assembly processes, highlighting the need for more practical, scalable, and interpretable solutions \cite{shyalika2024comprehensive,shyalika2024evaluating,mccormick2025real}

%Despite the advantages of multimodal anomaly prediction, several gaps remain in its application to manufacturing assembly processes. One significant challenge is data scarcity, as comprehensive datasets representing the entire manufacturing environment are often difficult to obtain. Additionally, existing methods struggle with poor performance metrics, especially in accurately predicting anomalies ahead of time in highly dynamic and complex assembly processes. The integration of diverse data sources often poses challenges in ensuring real-time processing and maintaining data consistency. Furthermore, the interpretability of anomaly predictions is limited, making it difficult for operators to understand and act on the predictions. Existing works have not fully addressed these issues, leaving a gap in practical, scalable, and interpretable solutions for multimodal anomaly prediction in manufacturing settings.

Multimodal fusion has emerged recently as a promising approach for integrating multiple data modalities to enhance the accuracy and robustness of various applications \cite{zhang2024multimodal,jaafar2023multimodal}. This integration leverages the complementary strengths of each modality, enabling a more comprehensive analysis. Decision-level fusion (a.k.a. late fusion) is a fusion technique that combines the decisions of multiple classifiers into a shared decision about any activity that occurred \cite{roggen2013signal}. Here, each classifier independently analyzes features and makes preliminary decisions, which are then aggregated into a fused decision vector to derive the final task decision\cite{kaya2017video,meng2013depression}. However, since the classifiers operate independently, decision-level fusion often fails to capture implicit semantic associations among the variables, leading to a loss of context, inconsistencies among classifier outputs, and challenges in interpretability and scalability.

%However, decision-level fusion can result in the loss of detailed context-related information, inconsistencies among classifier outputs, and challenges in interpretability and scalability, as it combines decisions from multiple classifiers without fully integrating or understanding the underlying data features. 

%A neurosymbolic approach can address these issues by preserving detailed semantic information, harmonizing and refining classifier outputs, and enhancing interpretability through explicit reasoning mechanisms. This approach facilitates scalable integration of multiple modalities through flexible adaptation. Additionally, incorporating transfer learning\cite{torrey2010transfer, pan2009survey}, which leverages pre-trained models, can further enhance predictive modeling by improving generalizability and efficiency, serving as a catalyst in advancing predictive capabilities.
Neurosymbolic AI integrates neural network-based techniques with symbolic knowledge-based approaches, offering two key perspectives for understanding this integration: (1) algorithmic-level considerations and (2) application-level considerations in AI systems \cite{Sheth2023NeurosymbolicA}. 
%In this work, we use a knowledge infusion approach to Neurosymbolic AI. 
This work uses a neurosymbolic AI-based knowledge-infusion for anomaly prediction. Knowledge-infusion allows us to integrate domain knowledge with data-driven approaches, providing a foundation for more robust and interpretable AI models. Building on this concept, we propose NSF-MAP (\underline{N}euro\underline{s}ymbolic \underline{M}ultimodal \underline{F}usion for Robust and Interpretable \underline{A}nomaly \underline{P}rediction), a novel framework specifically designed for robust and interpretable anomaly prediction in manufacturing pipelines. Our main contributions are:

\begin{itemize}
\item A novel anomaly prediction model for assembly pipelines that leverages neurosymbolic AI-based decision-level fusion combined with transfer learning.
\item Two derived datasets, one multimodal and one analog, designed for anomaly prediction in assembly pipelines, developed using domain-specific insights.
\item Integration of advanced user-level explainability techniques, including an ontology-based method, to enhance the interpretability of our predictions.
\end{itemize}

%\noindent The dataset, codes to reproduce the results and supplementary materials are available at  \url{https://anonymous.4open.science/status/NS-HyMAP-1EB9}.

\section{Related Work}

\subsection{Time Series and Image Fusion Learning Models}
\vspace{-0.5 mm}

Significant advancements in time series and image fusion models have been made across various domains. The ESTARFM model \cite{wang2014spatial} enhances spatial and temporal resolution by merging time series data with images. Hybrid models like SRCNN and LSTM \cite{yang2021robust} improve spatial details and temporal patterns in remote sensing. In healthcare, integrating electronic health records with radiographic images enhances the classification of atypical femur fractures \cite{schilcher2024fusion}. In manufacturing, the MFGAN model \cite{qu2024mfgan}, an attention-based autoencoder with a generative adversarial network, uses multimodal temporal data for anomaly prediction. \cite{iwana2020time} proposes a fusion approach combining DTW-derived features with CNNs for time series classification. Despite these advancements, multimodal fusion in manufacturing faces challenges such as data synchronization, computational demands, model complexity, scalability, real-time processing, data quality, and interpretability.

\vspace{-2 mm}
\subsection{Transfer Learning Techniques}
\vspace{-0.5 mm}
%Transfer learning has emerged as a pivotal approach in anomaly detection and anomaly prediction, enabling models to leverage knowledge from related tasks to improve performance with minimal additional labeled data \cite{torrey2010transfer, pan2009survey}. This technique addresses the challenge of limited datasets and dynamic environments, as seen in industrial processes\cite{yan2024comprehensive,maschler2021towards}, video surveillance \cite{jayaswal2021framework}, and web services \cite{zhang2022efficient}, by effectively transferring learned representations and adapting to new domains. Leveraging methods such as Variational Auto-Encoders, Graph Attention Networks, and fuzzy classifiers, transfer learning has enhanced the accuracy and efficiency of anomaly detection across diverse applications, from manufacturing to video anomaly detection \cite{series2024design,lughofer2022transfer}. The limitations of using transfer learning in anomaly detection and prediction include potential challenges with domain mismatch, reduced performance in significantly different target domains, and the necessity for careful adaptation to new tasks.

Transfer learning has emerged as a pivotal approach in anomaly detection and anomaly prediction, enabling models to leverage knowledge from related tasks to improve performance with minimal additional labeled data \cite{torrey2010transfer,pan2009survey}. It addresses issues with limited datasets and dynamic environments in fields like industrial processes \cite{yan2024comprehensive,maschler2021towards}, video surveillance \cite{jayaswal2021framework}, and web services \cite{zhang2022efficient}, by transferring learned representations and adapting to new domains. Techniques such as Variational Auto-Encoders, Graph Attention Networks, and fuzzy classifiers have improved anomaly detection accuracy and efficiency \cite{series2024design,lughofer2022transfer}. However, challenges include domain mismatch, reduced performance in divergent target domains, and the need for careful task adaptation.

%\vspace{-8mm}
\vspace{-2 mm}
\subsection{Neurosymbolic AI in Manufacturing}
\vspace{-0.5 mm}
Early research 
%has been conducted using knowledge graphs in anomaly detection \cite{Moghaddam2023AnomalyDF, Zhuo2021SurveyOS, Zhao2022ImprovingAD}. \cite{Zhuo2021SurveyOS, wang2023data} proposes an anomaly detection scheme leveraging a cybersecurity knowledge graph, utilizing graph neural networks to embed vertices and assess threat levels, ultimately ranking anomalies within the graph based on structural and attribute features. A knowledge graph has been developed by \cite{Zhao2022ImprovingAD} for power databases to evaluate anomaly detection algorithms and utilize Neo4j to enhance data processing visibility and anomaly traceability. \cite{inproceedings} proposes a hybrid neuro-symbolic method combining data-driven models and industrial ontologies for anomaly detection in Industry 4.0, while \cite{wang2023data} developed and validated a predictive maintenance method for industrial robots using data-driven models and knowledge graphs, enhancing intelligent manufacturing stability.
has explored knowledge graphs for anomaly detection \cite{Moghaddam2023AnomalyDF,Zhuo2021SurveyOS,Zhao2022ImprovingAD}. \cite{Zhuo2021SurveyOS,wang2023data} propose a scheme leveraging a cybersecurity knowledge graph with graph neural networks to embed vertices and assess threat levels, ranking anomalies based on structural and attribute features. \cite{Zhao2022ImprovingAD} uses a knowledge graph for power databases to evaluate detection algorithms and improve data visibility and traceability with Neo4j. \cite{inproceedings} introduces a hybrid neuro-symbolic method combining data-driven models and industrial ontologies for anomaly detection in Industry 4.0. \cite{wang2023data} presents a predictive maintenance approach for industrial robots using data-driven models and knowledge graphs, enhancing manufacturing stability.

%\cite{yan2022neuro} introduces Neuro-Symbolic Time Series Classification (NSTSC), a model combining signal temporal logic and neural networks to perform time series classification with interpretable outputs represented as STL formulas. Zhuo el al.

To the best of the authors' knowledge, no existing approach uses a neurosymbolic AI-based multimodal fusion for anomaly prediction in manufacturing contexts. A promising approach could involve integrating symbolic reasoning with neural networks based on multimodal data to enhance the interpretability and robustness of anomaly prediction models in complex industrial environments.

\begin{figure*}[!htb]
  \centering
  \vspace{-4 mm}\includegraphics[width=0.999\linewidth]{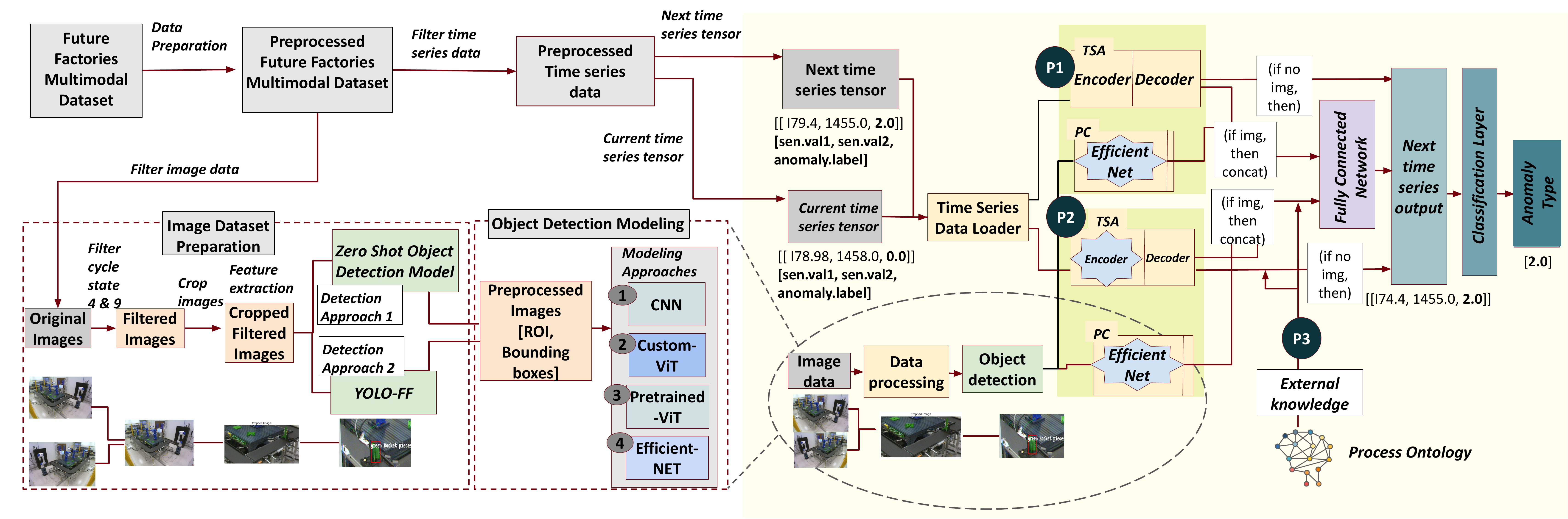}
\caption{Architecture of NSF-MAP: Integration of time series and images for anomaly prediction, involving preprocessing, feature extraction using a pretrained EfficientNet (PC), and fusion with time series autoencoder outputs(TSA). The fusion model, enhanced by external process ontology knowledge, predicts the next time series output and classifies anomaly types.}
  \label{fig:overall_architecture}
  \vspace{-4 mm}
\end{figure*}

\vspace{-2mm}
\section{Data Preparation}
\label{sec:dp}
\vspace{-1mm}
\paragraph{Time Series Data Preparation} We use the publicly available manufacturing dataset~\cite{harik2024analog} generated by the Future Factories (FF) Lab at the McNair Aerospace Research Center at the University of South Carolina. This dataset has two versions: analog and multimodal datasets. The dataset consists of measurements using a prototype of rocket assembly pipeline [see Appendix~\ref{sec:A1}], which adheres to industrial standards in deploying actuators, control mechanisms, and transducers.
%The multimodal dataset was generated from the assembly and disassembly of a rocket prototype. 
%The data includes sensor values of potentiometer and loadcell sensors, which are mounted onto the robots' end effectors. 
The multimodal dataset includes synchronized images captured from two cameras positioned on opposite sides of the testbed, which continuously record the operations. It consists of 166K records throughout the entire runtime of 30 hours with a sampling rate of 1.95 Hz. 
%The images are divided into batches of 1K samples of both camera viewpoints and stored in separate folders. For each batch, an equivalent JSON file containing the synchronized sensor values and their corresponding image paths can be found. Given the number of records, the dataset folder consists of a total of 166 image batch folders and their respective JSON files. In the data preparation, we first converted the data in JSON files into tabular format along with the path to the two camera images in the respective image batch folders. 
The images are divided into batches of 1K samples from both camera viewpoints, with each batch having an equivalent JSON file containing synchronized sensor values and their corresponding image paths. 
%The dataset consists of a total of 166 image batches and their respective JSON files. 
In data preparation, data from JSON files were first converted into a tabular format, including the paths to the camera images.
The data consists of 285 assembly cycles, at which one cycle represents a complete assembly and disassembly of the rocket. The rocket assembly process at the FF lab is divided into 21 distinct cycle states. Information about these cell cycle states and cycle state-wise anomaly types is not directly available in both the analog and multimodal datasets and we extracted them using a mapping function provided by domain experts. The preprocessed dataset includes measurements, such as sensor variables, conveyor variable frequency, drive temperatures, physical properties of robots (i.e., angles), conveyor workstation statistics, cycle state, cycle count, anomaly types and corresponding image file names from both cameras. In this paper, we introduce the preprocessed multimodal and analog datasets, incorporating this new information. Table 1 presents the statistics of anomalies in the preprocessed multimodal dataset used in the proposed method of this study.

%In this work, we first utilized, XGBoost 2.0.1,
%and its coverage measure to narrow down 20 out of the 41 measurements that contain high information content. XGBoost has achieved a SOTA performance on anomaly \textit{detection} and \textit{prediction} (prediction refers to the identification before the anomalous event, and detection refers to the identification after the event), and therefore, we used it to narrow down our feature selection (please refer to Appendix~\ref{sec:xgboost} for coverage plots and an example of a learned tree from the XGBoost model). Assembly cycles are associated with anomalies among seven different anomaly types. 
% \usepackage{tabularray}

% \usepackage{tabularray}
\begin{table}
\scriptsize
\centering
\caption{Anomaly types in the preprocessed multimodal dataset}
\label{tab:fil_anomalies}
\vspace{-2mm}
\begin{tblr}{
  cells = {c},
  cell{1}{2} = {c=2}{},
  cell{1}{4} = {c=2}{},
  vline{2-3} = {1}{},
  vline{2,4} = {2-10}{},
  hline{1-3,10-11} = {-}{},
}
                           & {Time series~\\Data} &            & Image Data &            \\
{Anomaly \\type}           & Count                & Percentage & Count      & Percentage \\
No Anomaly                 & 90844                & 54.72\%    & 10022      & 64.27\%    \\
NoBody1                    & 1849                 & 1.11\%     & 1312       & 8.41\%     \\
NoNose                     & 19307                & 11.63\%    & 1110       & 7.12\%     \\
NoNose,NoBody2             & 25206                & 15.18\%    & 1530       & 9.81\%     \\
{NoNose,NoBody2,\\NoBody1} & 26628                & 16.04\%    & 1620       & 10.39\%    \\
NoBody2                    & 1078                 & 0.65\%     & -          & -          \\
NoBody2,~NoBody1           & 1089                 & 0.65\%     & -          & -          \\
Total                      & 166001               &            &            &            
\end{tblr}
\vspace{-4mm}
\end{table}

%\subsection{Image Data Preparation}

%We use the derived image dataset from \cite{prasad2024assemai}. In this assembly pipeline, the rocket and its components are only visible during certain cycle states due to the positions of robots, machinery, and cameras. Based on domain expertise and observational insights, the images are filtered images from cycle four and part of cycle nine, which helps improve image analysis in the assembly process.
%After that, the filtered images are cropped to extract only the region of interest from the images. Appendix A illustrates the distribution of total image counts by anomaly labels in the preprocessed image dataset.

\paragraph{Image Data Preparation}
We use the derived image dataset\footnote{\url{https://github.com/renjithk4/AssemAI}} from \cite{prasad2024assemai}. In this assembly pipeline, rockets and components are visible only during certain cycle states due to the positions of robots, machinery, and cameras. Using domain expertise, images from cycle four and part of cycle nine are filtered to enhance analysis and further cropped to focus on the region of interest. %Appendix~\ref{appendixB}], shows the distribution of image counts by anomaly labels in the preprocessed dataset.

\vspace{-2mm}
\section{Problem Formulation}

Consider a multimodal dataset \( \mathcal{M} \), comprising time series data \( \mathcal{T} \) and  corresponding image data \( \mathcal{V} \), collected while assembling a rocket. The dataset consists of \( n \) cycles, denoted by \( \mathcal{C} = \{c_1, c_2, \ldots, c_n\} \), where each cycle \( c_i \) for \( i \in \{1, 2, \ldots, n\} \) represents the complete assembly of a product. Each cycle \( c_i \) is divided in to 21 distinct states, represented by \( \mathcal{S} = \{s_1, s_2, \ldots, s_{21}\} \). Let \( \mathcal{V}_{c_i, s_j} = \{v_{c_i, s_j, 1}, v_{c_i, s_j, 2}, \ldots, v_{c_i, s_j, T}\} \) denote the image data for cycle \( c_i \) at state \( s_j \).  \( \mathcal{T} \) comprises multivariate sensor readings at various time steps within each cycle state. Let \( \mathcal{T}_{c_i, s_j} = \{ \mathbf{t}_{c_i, s_j, 1}, \mathbf{t}_{c_i, s_j, 2}, \ldots, \mathbf{t}_{c_i, s_j, T} \} \), where \( \mathbf{t}_{c_i, s_j, k} = (t_{c_i, s_j, k, 1}, t_{c_i, s_j, k, 2}, \ldots, t_{c_i, s_j, k, m}, y_{c_i, s_j, k}) \) represents the multivariate sensor measurements at time step \( k \) for cycle \( c_i \) at state \( s_j \). Here, \( m \) is the number of sensor variables, \( t_{c_i, s_j, k, l} \) denotes the sensor measurement for the \( l \)-th sensor variable, and \( y_{c_i, s_j, k} \) is the anomaly label associated with time step \( k \). Additionally, let \( \mathcal{R} = \{R^{(1)}, R^{(2)}, \ldots, R^{(m)}\} \) denote set of acceptable sensor ranges, where \( R^{(l)} \) represents the acceptable range for the \( l \)-th sensor variable. The objective is to develop a model, \( \mathcal{F} \), for predicting  next sensor values \( \mathbf{t}_{c_i, s_j, k+1} \) and anomaly label \( y_{c_i, s_j, k} \) in the next time step of the assembly process.

%This involves integrating features from both image and time series data to enhance the predictive metrics for future sensor values and improve the accuracy of anomaly detection.

\section{NSF-MAP Method}
This section describes the NSF-MAP method (Figure \ref{fig:overall_architecture}), including feature extraction, fusion, and prediction approach. Then, we elaborate on the formulation of the loss function and the integration of the neurosymbolic AI approach utilized in this framework.

\subsubsection{Image and Time Series Feature Extraction}
We focus on images captured at states \( s_4 \) and \( s_9 \), which are critical for anomaly prediction. The region of interest(rocket parts) for each image \( V_{ci,sj} \) is identified using a bounding box \( B_{ci,sj} = (x_{\text{min}}, y_{\text{min}}, x_{\text{max}}, y_{\text{max}}) \), which is obtained from the YOLO-FF model \cite{prasad2024assemai}. Then, we use a pretrained EfficientNet model \( \mathcal{G}_V \) with the classification layer removed to extract the trained image embeddings  \( \mathbf{f}_V \) as defined in equation \ref{eq:second_equation}.

%This model is fine-tuned specifically for the images to focus on the rocket parts.
%Each image \( \mathcal{V}_{c_i, s_j} \) is preprocessed by cropping using a bounding box \( B_{c_i, s_j} = (x_{\text{min}}, y_{\text{min}}, x_{\text{max}}, y_{\text{max}}) \), obtained through YOLO-FF \cite{prasad2024assemai}, a object detection model finetuned to the images in FF multimodal dataset \(\mathcal{D}\) (defined in equation \ref{eq:first_equation}) to get the region of interest (rocket parts). 

%\vspace{-1mm}
%\begin{equation}
%\label{eq:first_equation}
%B_{c_i, s_j} = \mathcal{D}(\mathcal{V}_{c_i, s_j})
%\end{equation}
\vspace{-4mm}
\begin{equation}
\label{eq:second_equation}
\mathbf{f}_V = \mathcal{G}_V(\mathcal{V}_{c_i, s_j})
\end{equation}
\vspace{-4mm}

\noindent For time series data, an autoencoder-decoder network \( \mathcal{G}_T \) is used to extract features into a latent representation \( \mathbf{f}_T \) and to reconstruct the input sequence. The encoder \( \mathcal{E}_T \) (defined in equation \ref{eq:third_equation}) extracts features, while the decoder \( \mathcal{P}_T \) (defined in equation \ref{eq:fourth_equation}) reconstructs the sequence.
\begin{equation}
\label{eq:third_equation}
\mathbf{f}_T = \mathcal{E}_T(\mathcal{T}_{c_i, s_j})
\end{equation}
\vspace{-4mm}
\begin{equation}
\label{eq:fourth_equation}
\mathbf{h}_T = \mathcal{P}_T(\mathbf{f}_T)
\end{equation}
The overall process of the autoencoder-decoder network \( \mathcal{G}_T \) can be described as in equation \ref{eq:fifth_equation}.
\begin{equation}
\label{eq:fifth_equation}
\mathcal{G}_T(\mathcal{T}_{c_i, s_j}) = \mathcal{P}_T(\mathcal{E}_T(\mathcal{T}_{c_i, s_j})) = \mathbf{h}_T
\end{equation}

\subsubsection{Decision-level Fusion and Prediction}
In this approach, we integrate the features extracted from the models \( \mathcal{G}_T \) and \( \mathcal{G}_V \). The features \( \mathbf{f}_V \) from \( \mathcal{G}_V \) and \( \mathbf{h}_T \) from \( \mathcal{G}_T \) are concatenated to form a unified feature vector \( \mathbf{z} \) as in equation \ref{eq:sixth_equation}.
\begin{equation}
\label{eq:sixth_equation}
\mathbf{z} = [\mathbf{f}_V; \mathbf{h}_T]
\end{equation}

\noindent This unified feature vector \( \mathbf{z} \) is then passed through a fully connected network \( \mathcal{H} \) to obtain the final prediction as in equation \ref{eq:seventh_equation}.
\begin{equation}
\label{eq:seventh_equation}
\mathbf{y}_{c_i, s_j, k+1} = \mathcal{H}(\mathbf{z})
\end{equation}

where \( \mathbf{y}_{c_i, s_j, k+1} \) represents both the predicted sensor values \( \hat{\mathbf{t}}_{c_i, s_j, k+1} \) and the anomaly label \( \hat{y}_{c_i, s_j, k+1} \) at the next time step \( k+1 \) for cycle \( c_i \) at state \( s_j \). In all other states except for states \( s_4 \) and \( s_9 \), the features \( \mathbf{h}_T \) from the autoencoder-decoder network are used directly for prediction as defined in equation \ref{eq:eight_equation}.
\begin{equation}
\label{eq:eight_equation}
\mathbf{y}_{c_i, s_j, k+1} = \mathcal{H}(\mathbf{h}_T)
\end{equation}
We illustrate the decision-level fusion and prediction in Figure \ref{fig:overall_architecture} (Method P1).

\subsubsection{Limitations and Transfer Learning Strategy}
The above approach, where the entire autoencoder-decoder network is trained end-to-end, may face limitations due to the potential overfitting of the decoder to specific features of the time series data, which might not generalize well to unseen data. Training the entire network might also result in longer training times and increased computational costs. Transfer learning, which involves reusing a pre-trained model for a new problem \cite{torrey2010transfer,pan2009survey}, is utilized here by freezing the encoder of the autoencoder while training only the decoder. When the encoder \( \mathcal{E}_T \) is frozen, its parameters are not updated during the training of the decoder \( \mathcal{R}_T \). This means that only the decoder parameters \( \theta_{\mathcal{R}_T} \) are optimized. In this scenario, the prediction model \( \mathcal{H} \) uses the features \( \mathbf{h}_T \) extracted by the fixed encoder and the trained decoder as in equation \ref{eq:eight_equation}. The updated objective function, incorporating the model freezing, remains the same as previously described, but now \( \mathbf{f}_T \) is generated with the encoder \( \mathcal{E}_T \) held constant. We illustrate the incorporation of transfer learning in Figure \ref{fig:overall_architecture} (Method P2).
%\[
%\mathbf{f}_T = \mathcal{E}_T(\mathcal{T}_{c_i, s_j}) \quad \text{(Encoder is frozen)}
%\]
%\[
%\mathbf{h}_T = \mathcal{D}_T(\mathbf{f}_T) \quad \text{(Decoder is %trained)}
%\]

%\vspace{-2 mm}
%\subsection{Knowledge-Infusion and Explainability}
%\label{sec:KI}
%\subsubsection{Dynamic Process Ontology} \label{sec:process_ontology}
\subsubsection{Process Ontology for Knowledge-Infusion} 
We design and develop Dynamic Process Ontology\footnote{\url{https://github.com/revathyramanan/Dynamic-Process-Ontology}} to support knowledge-infused learning and provide user-level explanations for model predictions. The ontology is built on the basis of 21 cycle states \( \mathcal{S} = \{s_1, s_2, \ldots, s_{21}\} \), capturing the temporal component of the assembly process and hence the name \textit{Process Ontology}. Additionally, the ontology can be \textit{dynamically} updated as per the experimentation set-up to adjust the expected minimum and maximum values of the sensor, the function of robots in the cycle states, and other information as required (example in Appendix~\ref{appendixC}). The specific attributes of process ontology are as follows: (a) definitions and item specifications of each sensor and equipment, (b) relationships between the sensors and equipment, (c) function and involvement of each sensor and robot with respect to the cycle states, (d) expected (or anomalous) values of sensor variables concerning each cycle state, (e) types of anomalies that could be associated with each cycle state, and (f) sensor values and properties that can be dynamically updated as per changes in the experimental set-up.  

While each sensor has possible minimum and maximum values that remain constant throughout the assembly process, the expected normal operating range varies based on the sensor's role in a given cycle state. For example, a potentiometer's expected range on robot-1 varies by cycle state. This necessitates an understanding of the relationships between cycle states, robot involvement, and corresponding sensor readings. Therefore, we gather this domain knowledge from the Process Ontology to guide the model training process. An ontology dynamically integrates sensor ranges by adapting expected values to cycle states, workload, and equipment interactions, enabling real-time reasoning and explainability beyond static thresholds. Furthermore, in this work, we integrated domain-specific knowledge, particularly sensor ranges \( \mathcal{R} = \{R^{(1)}, R^{(2)}, \ldots, R^{(m)}\} \), to improve the model's performance and enhance its robustness. By incorporating domain constraints into the learning process, the Knowledge Infusion approach ensures that the model generalizes beyond fixed thresholding, reduces false positives and negatives, and effectively reasons over complex multi-sensor dependencies that cannot be captured through simple thresholding alone.

In model training, the predictions must satisfy:
\[
R^{(l)}_{\min} \leq \mathbf{y}_{c_i, s_j, k+1}^{(l)} \leq R^{(l)}_{\max}, \quad \forall l = 1, 2, \ldots, m
\]
%\vspace{-1 mm}
The penalty \( P \) is added in two specific scenarios, as outlined in Equation \ref{eq:twelve_equation}. A penalty is imposed when the model predicts an anomaly even though the sensor values are within the acceptable range, indicating that the model incorrectly identifies an anomaly where none exists. A penalty is also applied when the model predicts normal conditions despite the sensor values being outside the acceptable range, reflecting a failure to detect an anomaly. 
\begin{equation}
\label{eq:twelve_equation}
\begin{aligned}
P = & \sum_{i=1}^{n} \sum_{j=1}^{21} \sum_{l=1}^{m} \left[ \mathbf{1}_{\left\{ R^{(l)}_{\min} \leq \hat{\mathbf{t}}_{c_i, s_j, k+1}^{(l)} \leq R^{(l)}_{\max} \right\}} \cdot \mathbf{1}_{\{\text{anomaly}\}} \right. \\
& \left. + \mathbf{1}_{\left\{ \hat{\mathbf{t}}_{c_i, s_j, k+1}^{(l)} < R^{(l)}_{\min} \, \text{or} \, \hat{\mathbf{t}}_{c_i, s_j, k+1}^{(l)} > R^{(l)}_{\max} \right\}} \cdot \mathbf{1}_{\{\text{normal}\}} \right]
\end{aligned}
\end{equation}

\noindent where \( \mathbf{1} \) is an indicator function that adds a penalty when the conditions inside it are met.

\subsubsection{Loss Function}
We employ a Weighted Mean-Squared Error (WMSE) loss function to address the class imbalance and  to further differentiate between the different types of anomalies. Let \( \mathbf{w} \) be the vector of class weights, \( \mathbf{y} \) be the true values, and \( \mathbf{{p}} \) be the predicted values, the loss function can be defined as in equation \ref{eq:ninth_equation}.
\vspace{-1 mm}
\begin{equation}
\label{eq:ninth_equation}
\mathcal{L}_{\text{wmse}} = \sum_{i=1}^{n} w_i \cdot (y_i - {p}_i)^2
\end{equation}

\noindent The objective is to minimize the WMSE for predicting the next sensor values and anomaly label as in equation \ref{eq:tenth_equation}.
\begin{equation}
\label{eq:tenth_equation}
\min_{\theta} \frac{1}{n} \sum_{i=1}^{n} \sum_{j \in \{4, 9\}} \sum_{k=1}^{T} \mathcal{L}_{\text{wmse}}(\mathbf{y}_{c_i, s_j, k+1}, 
\mathbf{y}_{c_i, s_j, k})
\end{equation}

\noindent The final predictions \( \mathbf{y}_{c_i, s_j, k+1} \) are then fed into a classification layer \( \mathcal{C} \), which will filter and classify the anomaly label \( \hat{y}_{c_i, s_j, k+1} \) into separate anomaly types (equation \ref{eq:elevnth_equation}).
\vspace{-0.25 mm}
\begin{equation}
\vspace{-0.25 mm}
\label{eq:elevnth_equation}
\hat{a}_{c_i, s_j, k+1} = \mathcal{C}(\mathbf{y}_{c_i, s_j, k+1})
\end{equation}

where \( \hat{a}_{c_i, s_j, k+1} \) represents the predicted anomaly type at the next time step \( k+1 \) for cycle \( c_i \) at state \( s_j \). The full objective function, with the penalty, is as in equation \ref{eq:thirteen_equation}.
\vspace{-1 mm}
\begin{equation}
\label{eq:thirteen_equation}
L=\frac{1}{c} \sum_{i=1}^{c} \frac{1}{s} \sum_{j=1}^{s} \frac{1}{k} \sum_{l=1}^{k} w_{i, j, l} \left( \mathbf{y}_{c_i, s_j, k+1}^{(l)} - \mathbf{y}_{c_i, s_j, k}^{(l)} \right)^2 + \lambda P
\end{equation}

\noindent where \( w_{i, j, l} \) are the sample weights, \( \lambda \) is a hyperparameter controlling the strength of the penalty term and \( P \) is the penalty function defined above. We illustrate the knowledge-infusion approach in Figure \ref{fig:overall_architecture} (Method P3). The application of this ontology for providing user-level explanations is elaborated in the Results section.

%\begin{figure}[!htb]
%\centering{\includegraphics[width=\linewidth]%{figs/overall_architecture_DLF.pdf}}
%  \caption{Proposed Methods}\label{fig:overall_architecture_DLF}
 % \vspace{-5 mm}
%\end{figure}

\section{Experiments}
%We delineate the experimental configurations aimed at assessing NS-HyMAP's comprehensive performance and evaluating the individual contributions of its sub-components.

\subsection{Common Hyperparameters and Training Setup} 
\vspace{-0.5 mm}
%The preprocessed dataset is divided into 80\% for training and 20\% for testing. We use cycle-wise splitting to ensure that 80\% of cycles are used for training and 20\% for testing while maintaining the same percentage of normal and anomalous samples in both datasets. 
%We did above for the same by splitting the dataset in to  60\% for training and 40\% for testing, We use cycle-wise splitting to ensure that 60\% of cycles are used for training and 40\% for testing while maintaining the same percentage of normal and anomalous samples in both datasets. 
The dataset is split using a cycle-wise approach, ensuring 80-20\% and 60-40\% training-testing splits while maintaining the same proportion of normal and anomalous samples in both training and testing sets. For all models, we use WMSE, tuning hyperparameters like epochs(50), batch size(32), and learning rate(0.001). The best model is saved based on validation accuracy. All the model parameters are optimized using the Adam optimizer, and the training process involves a training phase to adjust the model weights and a validation phase to monitor the model's performance on unseen data. During preprocessing, we resize the original images to 224×224 pixels and normalize them using the mean [0.485,0.456,0.406] and standard deviation [0.229,0.224,0.225]. In the following section, we describe the experimentation details of baselines in B1 and B2 and proposed methods in P1, P2, and P3, respectively.

%\subsection{Baseline Methods}
\subsubsection{B1:Time Series-based Autoencoder Model}
%The model architecture used in the time-series-only model is a simple autoencoder-based regressor designed to predict the next time series values of sensor data and anomaly labels. 
The autoencoder model consists of two main components: an encoder and a decoder. The encoder is a single linear layer that compresses the input data (we use two most important sensor variables and anomaly label) into a lower-dimensional representation with a hidden size of 10, with a ReLU activation function applied to introduce non-linearity. The decoder is a single linear layer that reconstructs the original input from this compressed representation. The autoencoder aims to learn a compact representation of the input data that can effectively capture its essential features. The model optimizes the WMSE between the predicted next-time series values (sensor values and anomaly label) and the current values. 

\subsubsection{B2:Image-based EfficientNet-B0 model}
We implement an EfficientNet-B0 model \cite{tan2019efficientnet} on the preprocessed images for an anomaly detection task, initially pre-trained on the ImageNet dataset. We adapt the model by modifying the final classification layer to align with the number of classes in our image dataset (five classes). 
%The best model was saved based on validation accuracy.

%\subsection{Proposed Methods}
\subsubsection{P1: Decision Level Fusion}\textbf{(DLF)}
\label{sec:dlf}
(Figure \ref{fig:overall_architecture}: P1)
The autoencoder model is designed with an encoder that takes an input dimension of 3, a hidden dimension of 64, and a latent dimension of 128. It utilizes the ReLU activation function. The decoder receives the latent dimension of 128 and reconstructs the subsequent time series data with an output dimension of 3, also employing the ReLU activation function. The EfficientNet-B0 model is pre-trained and modified by removing its final classification layer. This alteration allows its features to be concatenated with the latent features from the time series before being passed to a fully connected layer with an output dimension of 3. The concatenated output of autoencoder and  EfficientNet-B0 is fed to a fully connected layer, which has a dropout rate of 0.5 for regularization. The model aims to predict the next time series data with an output dimension of 3.  A ReduceLROnPlateau scheduler adjusts the learning rate based on validation loss, with a patience of 5 epochs. Early stopping is implemented to prevent overfitting, with the best model saved based on the lowest validation loss.

\subsubsection{P2: Decision Level Fusion with Transfer Learning}\textbf{(DLF+TL)}(Figure \ref{fig:overall_architecture}: P2) 
The hyperparameters of autoencoder and EfficientNet-B0, training process, and loss function will be the same as in the P1. The encoder is frozen to disable gradient updates for the encoder parameters. 
%We keep the decoder part of the autoencoder trainable to adjust to any changes in the training data and to improve the reconstruction of the time series data from the latent representations.
%By keeping the decoder trainable, we allow the model to fine-tune its ability to reconstruct the input data. accurately. Keeping the decoder trainable helps avoid overfittingand adjusting the model to the specificities of the newtraining process, especially when integrating with additional modalities like images.

\subsubsection{P3: Enhanced Decision-Level Fusion with Transfer Learning through Neurosymbolic AI}\textbf{(DLF+TL+KIL)} (Figure \ref{fig:overall_architecture}: P3) 
The hyperparameters, training process, and loss function for the autoencoder and EfficientNet-B0 remain consistent with those in P1. In this proposed approach, we implement a custom loss function to P2 that includes WMSE loss with an additional penalty, as explained in the Methodology section. 
\vspace{-2mm}
\section{Results}

\begin{table*}
\scriptsize
\centering
\begin{minipage}{0.6\textwidth}
\caption{Experimental Results of NSF-MAP and Ablation Studies (mean(\%) ±  std(\%)). \textbf{D: model used only for Detection.} Number of test samples used in 80\% 20\% split: 33201, in 60\% 40\% split: 66402. \textbf{Bold} indicates the highest performance.}
\label{tab:results}
\begin{tblr}{
  row{odd} = {c},
  row{4} = {c},
  row{6} = {c},
  row{8} = {c},
  row{10} = {c},
  cell{2}{1} = {c},
  vline{2} = {-}{},
  hline{1-3,7-8,12} = {-}{},
}
                           & B1                                        & B2: *D                                                                       & {DLF \\(P1)}                              & {DLF (TS+\\zero tensor \\image)} & {DLF+TL \\(P2)}   & {DLF+\\KIL}      & {DLF+TL+\\KIL(P3)}                  \\
\textbf{\textbf{}}         & \textbf{\textbf{\textbf{\textbf{80\%~}}}} & \textbf{\textbf{\textbf{\textbf{\textbf{\textbf{\textbf{\textbf{20\%}}}}}}}} & \textbf{Split}                            &                                  &                   &                  &                                     \\
{Weighted Avg.\\Precision} & {74.00±\\1.00\%}                          & {97.00± \\1.05\%}                                                            & {76.05± \\2.00\%}                         & {80.00±\\1.00\%}                 & {91.00± \\0.05\%} & {93.00±\\1.00\%} & {\textbf{94.00± }\\\textbf{1.00\%}} \\
{Weighted Avg.\\Recall}    & {63.00±\\1.00\%}                          & {97.00±\\0.5\%}                                                              & {72.00±\\1.70\%}                          & {64.00±\\0.5\%}                  & {88.02±\\0.05\%}  & {90.00±\\0.05\%} & {\textbf{93.00±}\\\textbf{0.75\%}}  \\
{Weighted Avg.\\F1-Socre}  & {61.00±\\1.00\%}                          & {97.00±\\1.00\%}                                                             & {72.03±\\2.05\%}                          & {64.00±\\0.5\%}                  & {89.00±\\0.05\%}  & {91.00±\\0.05\%} & {\textbf{93.00±}\\\textbf{0.05\%}}  \\
{Weighted Avg.\\Accuracy}  & {63.00±\\1.00\%}                          & {97.00±\\1.00\%}                                                             & {72.00±\\2.00\%}                          & {64.00±\\0.05\%}                 & {88.00±\\0.05\%}  & {90.00±\\0.05\%} & {\textbf{93.00±}\\\textbf{1.00\%}}  \\
\textbf{}                  & \textbf{\textbf{60\%~}}                   & \textbf{\textbf{40\%}}                                                       & \textbf{\textbf{\textbf{\textbf{Split}}}} &                                  &                   &                  &                                     \\
{Weighted Avg.\\Precision} & {64.00±\\1.00\%}                          & {87.00±\\1.05\%}                                                             & {66.05±\\2.00\%}                          & {70.00±\\1.00\%}                 & {81.00±\\0.05\%}  & {83.00±\\1.00\%} & {\textbf{84.00±}\\\textbf{1.00\%}}  \\
{Weighted Avg.\\Recall}    & {53.00±\\1.00\%}                          & {87.00±\\0.5\%}                                                              & {62.00±\\1.70\%}                          & {60.00±\\0.5\%}                  & {78.02±\\0.05\%}  & {80.00±\\0.05\%} & {\textbf{83.00±}\\\textbf{0.75\%}}  \\
{Weighted Avg.\\F1-Socre}  & {51.00±\\1.00\%}                          & {87.00±\\1.00\%}                                                             & {64.00±\\0.5\%}                           & {58.00±\\0.5\%}                  & {79.00±\\0.05\%}  & {81.00±\\0.05\%} & {\textbf{83.00±}\\\textbf{0.05\%}}  \\
{Weighted Avg.\\Accuracy}  & {53.00±\\1.00\%}                          & {87.00±\\1.00\%}                                                             & {64.00±\\0.05\%}                          & {58.00±\\0.05\%}                 & {78.00±\\0.05\%}  & {80.00±\\0.05\%} & {\textbf{83.00±}\\\textbf{1.00\%}}  
\end{tblr}
\end{minipage}
    \hfill
\begin{minipage}{0.37\textwidth}
        \centering
        \includegraphics[width=\linewidth]{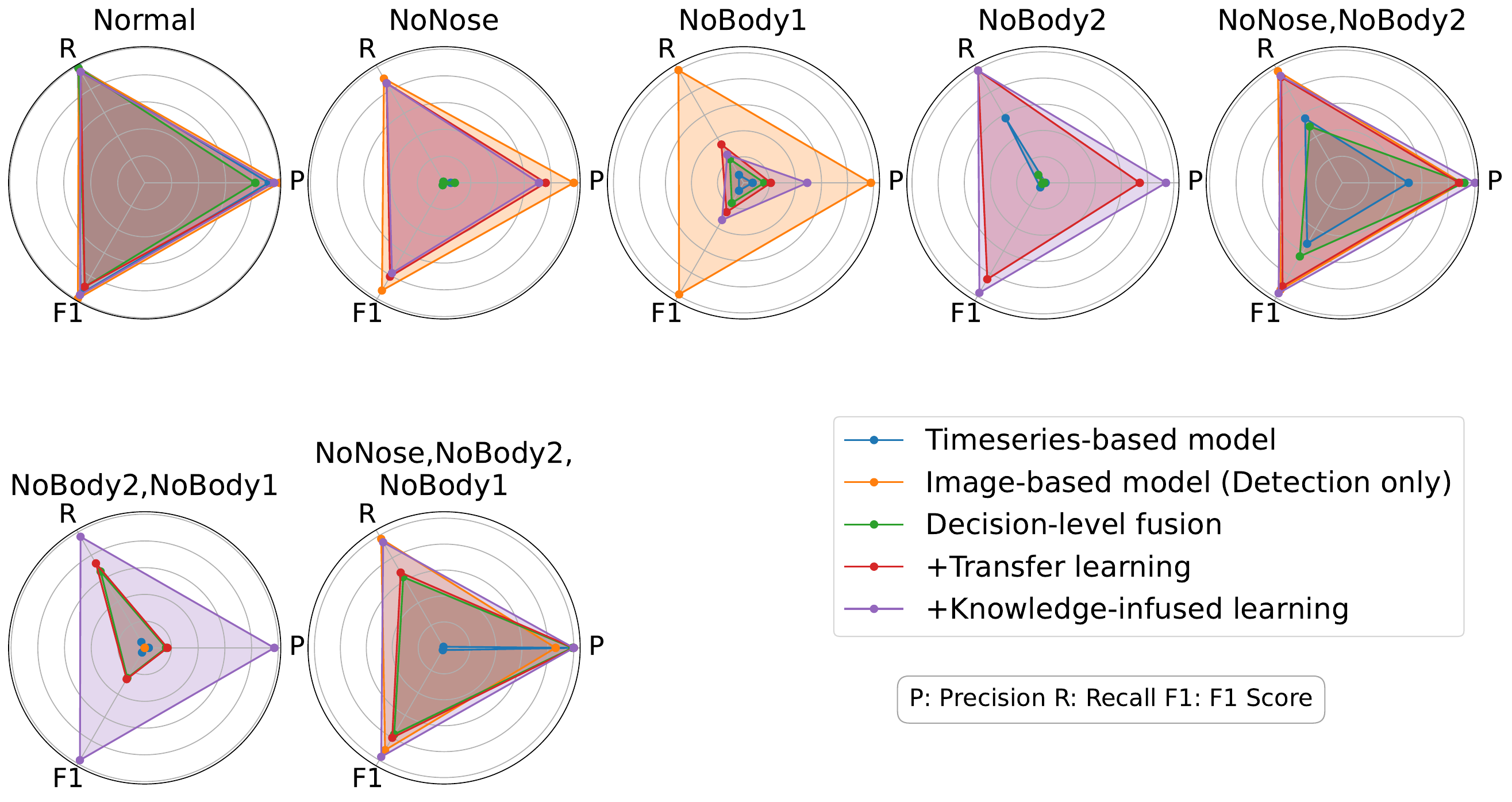}
        \captionof{figure}{Experimental Results of Predicting Different Anomaly Types by NSF-MAP}
        \label{fig:Experimental_Results}
        
        \includegraphics[width=\linewidth]{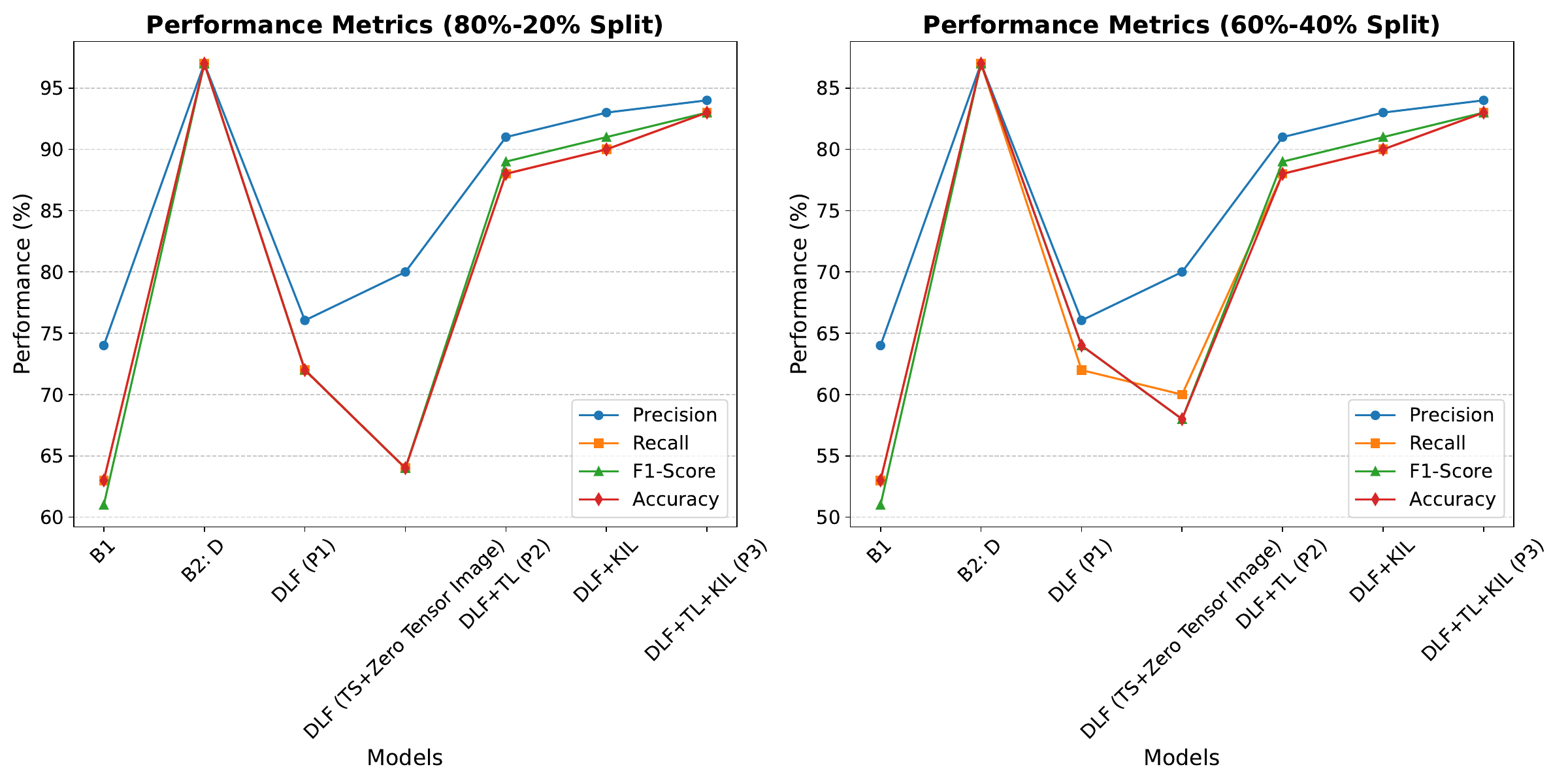}
        \captionof{figure}{Performance of the Proposed Approach and Baselines with Varied Training and Testing Splits}
        \label{fig:Experimental_Results2}
\end{minipage}
\end{table*}

Table 2 summarizes the results of our experiments on the test set and the ablation studies across different baselines. We evaluate the performance using four metrics: weighted averages of precision, recall, F1-score, and accuracy. The weighted averages are calculated based on the six types of anomaly classes and the normal class. In the 80-20\% split, the Decision-Level Fusion model achieves an overall accuracy of 72\%, with a precision of 76.05\%, a recall of 72\%, and an F1-score of 72.03\%. With the inclusion of Transfer learning module, the accuracy and the F1 score increased by 39\% and 45\%, respectively. Similarly, infusion of knowledge on sensors enhanced the accuracy and the F1 score by 47\% and 52\%, respectively. Figure \ref{fig:Experimental_Results} depicts the performance of detecting various anomaly types and the normal class across various modeling approaches. It can be observed that among all the models, Decision level fusion with transfer learning and knowledge-infused learning approach gives the best results in detecting five out of seven types of classes. Figure \ref{fig:Experimental_Results2}  presents the comparison of performance metrics for different models under the two training-testing split scenarios (80\%-20\% and 60\%-40\%), demonstrating the effectiveness of the proposed approach against baseline models.
\vspace{-3mm}
\subsection{Process Ontology for User-level Explainability}
\vspace{-4 mm}
Dynamic Process Ontology, introduced in the Methodology section, is utilized to provide user-level explanations for the outputs generated by the model. The model predicts the values of the sensors before passing them on to the classification layer. In both scenarios, whether an anomaly is present or not, the variable values predicted by the models can be used to explain the anomaly predicted by the model. 
For example, if a value for variable-1 is predicted to be 3000 in cycle state 4, where its expected values are supposed to be 6000 to 8000, the anomaly can be explained as follows: (i) which variable is responsible for the anomaly? (ii) what were the function of the robots during that state? (iii) what are the expected values of that variable in that state? The ontology can also capture misclassification in certain scenarios. If the model predicted the anomaly type to be "NoNose" in cycle state 4, it can be observed from ontology that "NoNose" anomaly happens only from cycle state 8. Similarly, if all the variables are predicted to be within the expected range for a given cycle state and the model predicted still predicted them as anomaly; these misclassifications can be identified by the ontology. To summarize, the ontology serves as a knowledge-infusion and an explainability tool along with serving as a mechanism to identify misclassifications in certain cases if at all those were missed by the model.

\vspace{-2 mm}
\section{Deployment of NSF-MAP}
We deploy the NSF-MAP and process ontology at the FF testbed at the McNair Center, University of South Carolina. The deployment strategy of this model consists of two main parts (See deployment setup in Appendix~\ref{appendixD}]). First, it requires establishing connections between the trained model, ontology, and the system. This includes real-time data retrieval through an Open Platform Communications Unified Architecture (OPC-UA) server on the system's Programmable Logic Controller (PLC) and accessing analog data via an OPC client script. Additionally, cameras around the system provide necessary images, accessed using custom Python libraries, \textit{Pyplon} and \textit{ImageCap}. Second, the internal pipeline is created to integrate the ontology and NSF-MAP. Data from the OPC-UA server is ingested into a process ontology, running on a local Neo4j server via Docker, to generate user-level explanations. These explanations and the corresponding images and analog data are then fed into NSF-MAP for real-time anomaly prediction. We face several challenges in deployment, which have been explained in Appendix~\ref{appendixD}].

\vspace{-2 mm}
\section{Conclusion and Further Work}
In this study, we derived industry-standard datasets tailored for analog and multimodal assembly processes. We developed a novel methodology for robust and interpretable anomaly prediction using neurosymbolic multimodal fusion. 
%A key aspect of our approach was enhancing interpretability by linking user-level explainability to high-level phenomena such as structural integrity, thereby improving model understanding and providing actionable insights for anomaly prediction. 
Our findings indicate that the time-series-image fusion model incorporated with a neurosymbolic AI approach offers significant improvements over traditional methods. Future research should investigate hybrid architectures that incorporate additional modalities, such as textual data, alongside advanced interpretability techniques for real-time anomaly prediction, enabling broader applicability across domains and production environments. Additionally, we aim to integrate images into the process ontology, capturing key timestamps using techniques like dynamic time warping, change-point detection, or saliency-based methods. Also, to improve how domain experts understand our approach, we suggest creating abstract representations of causal factors associated with anomalies, providing a more comprehensive view of anomalous events.

%\section*{Acknowledgment}
%Redacted for anonymity.
\vspace{-2 mm}
\section{Appendix}
\subsection{Appendix A. Future Factories Setup}
\label{sec:A1}
\begin{figure}[!htb]
  \centering
   \includegraphics[width=0.5\linewidth]{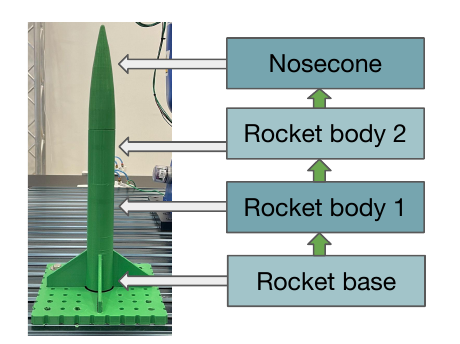}
   \vspace{-2mm}
  \caption{A Rocket Assembled by the Future Factories Lab. Any missing part is considered an anomaly: for example, the absence of Rocket body 1 is labeled as "NoBody1," while the absence of both Rocket body 1 and body 2 is labeled as "NoBody2, NoBody1."}
 \label{fig:rocket}
 \vspace{-2mm}
\end{figure}

\begin{figure}[!htb]
  \centering
   \includegraphics[width=0.8\linewidth]{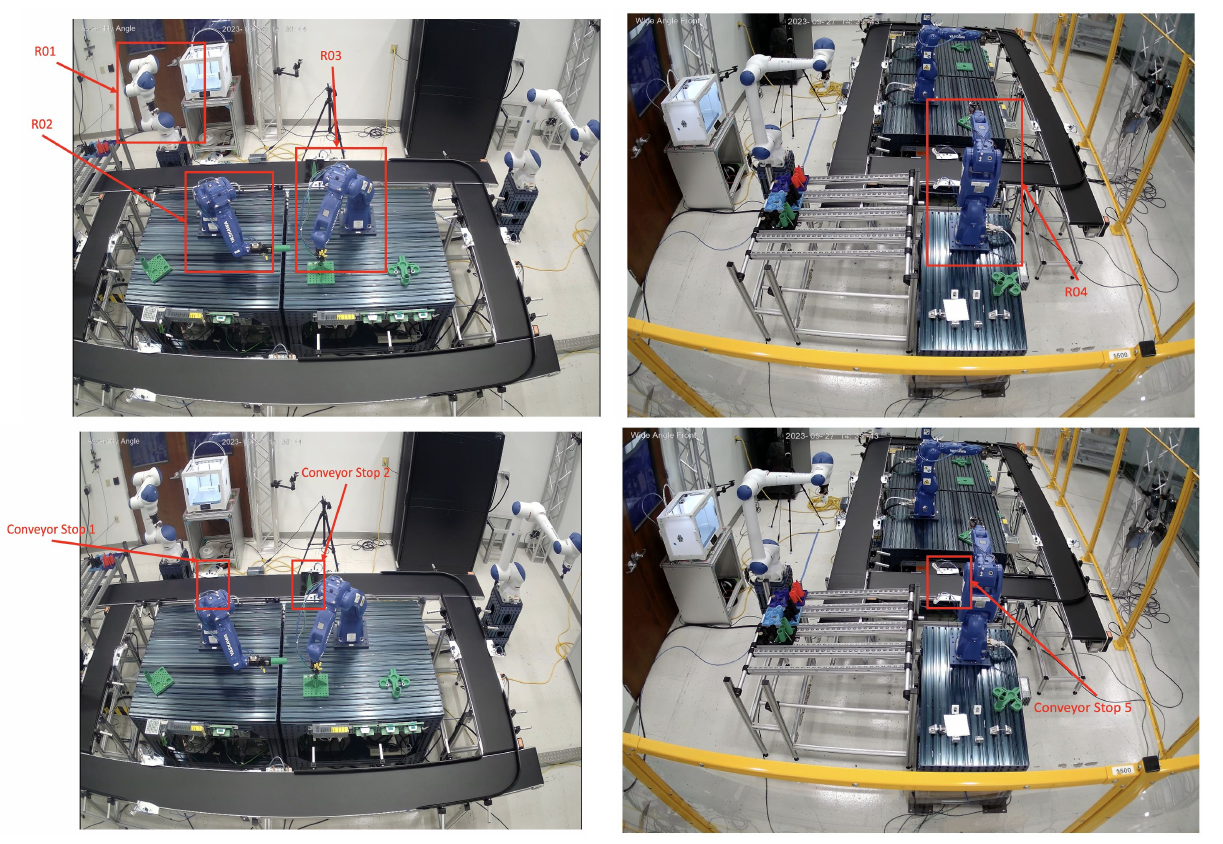}
   \vspace{-2mm}
  \caption{Future Factories Setup at Future Factories Lab. R01-Robot 1, R02-Robot 2, R03-Robot 3, R04-Robot 4}
  \label{fig:ffcell2}
\end{figure}

Figure \ref{fig:rocket} includes an image of a rocket assembled by FF Lab. A visual representation of the lab setup is shown in Figure \ref{fig:ffcell2}.
\vspace{-2mm}
\vspace{-2mm}

\vspace{-2mm}

%\label{appendix:anomaly_distribution}
%\begin{figure}[!htb]
 % \centering
  %\vspace{-4 mm}
   %\includegraphics[width=0.995\linewidth]{figs/Anomaly_Labels_Distribution.pdf}
  %\caption{Distribution of anomaly labels in the image dataset}
  %\label{fig:Anomaly_labels}
  %\vspace{-4mm}
%\end{figure}
\vspace{-4mm}
\subsection{Appendix B. Experiments with the Process Ontology}\label{appendixC}
\begin{figure}[!htb]
  \centering
  % adjust width to adjust the figure size
   \includegraphics[width=0.25\textwidth]{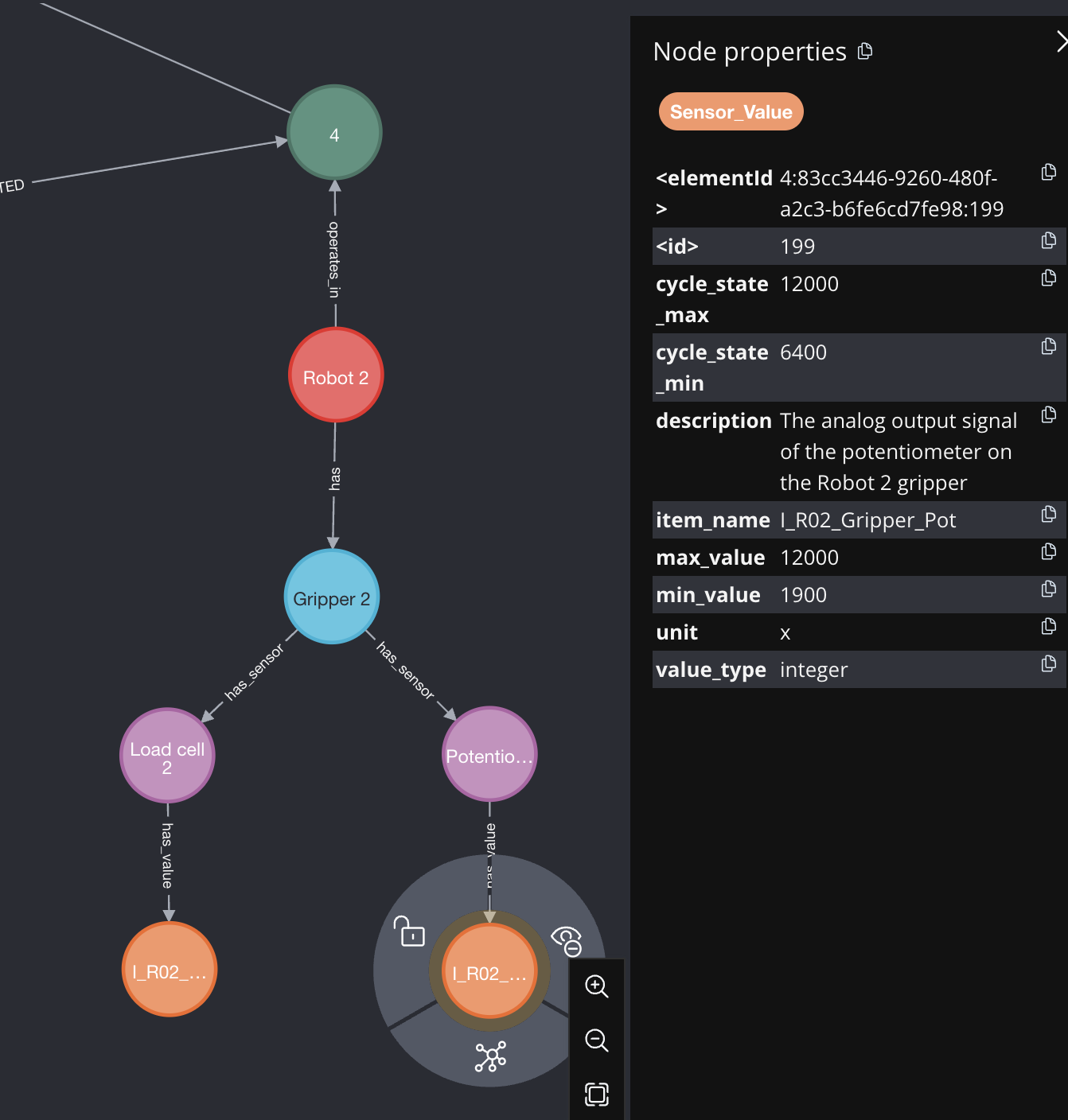}
   \vspace{-2mm}
  \caption{A Snapshot of Dynamic Process Ontology showcasing the properties of potentiometer}
  \label{fig:process_ontology}
\end{figure}

A snapshot of the ontology for a given cycle state is given in Figure \ref{fig:process_ontology}.

\vspace{-2mm}

\subsection{Appendix C. Deployment of NSF-MAP and Challenges Faced}\label{appendixD}
\begin{figure}[!htb]
\vspace{-4mm}
  \centering
   \includegraphics[width=\linewidth]{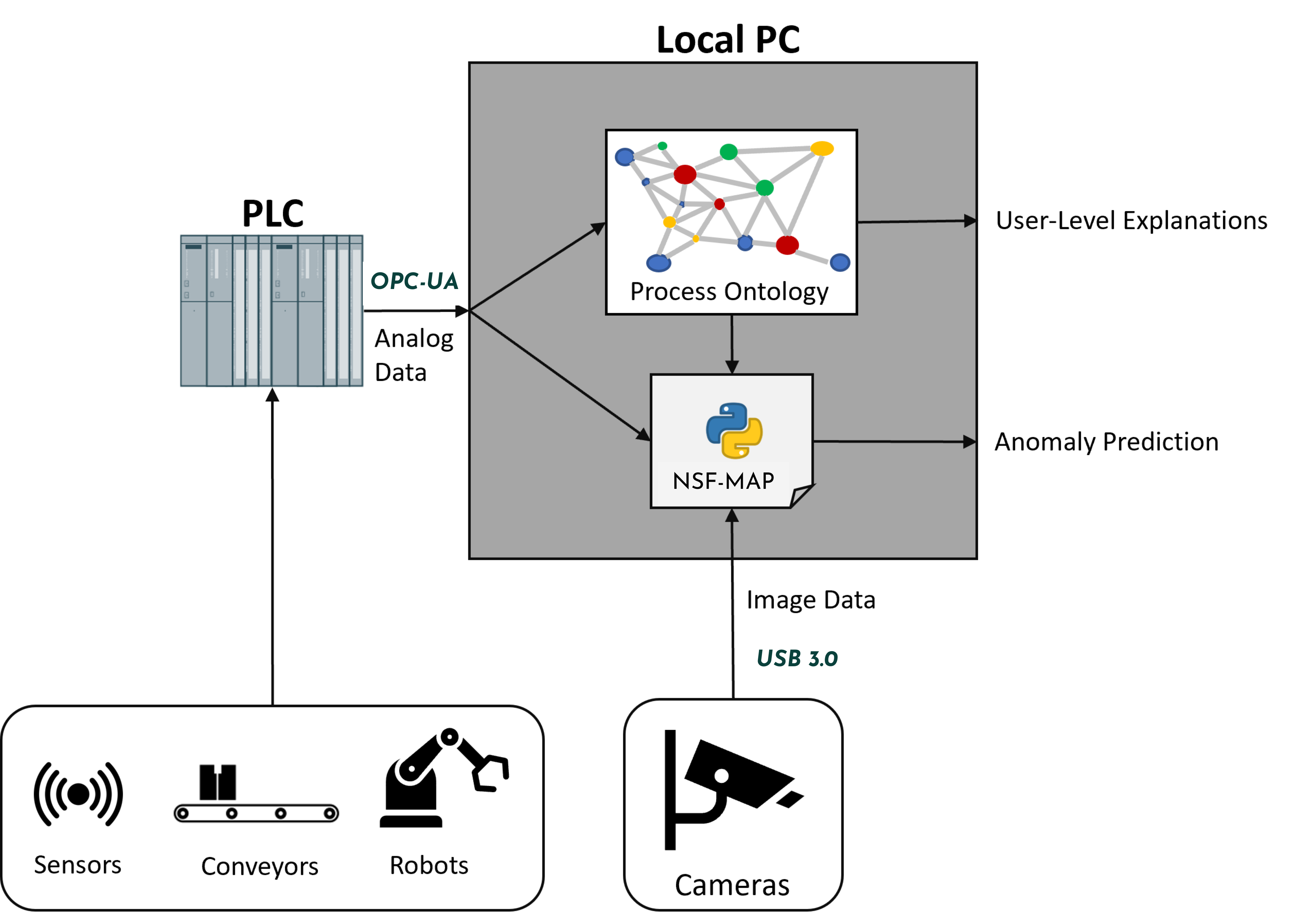}
   \vspace{-2mm}
  \caption{Deployment Setup of NSF-MAP}
  \label{fig:deployment_arc}
  \vspace{-6mm}
\end{figure}

%\subsubsection{Deployment Setup}
Figure \ref{fig:deployment_arc} illustrates the architecture for real-time deployment of the NSF-MAP model and process ontology onto the manufacturing system. It highlights integrating the trained model with the OPC-UA server for data retrieval and the connection to cameras for image acquisition, enabling seamless real-time predictions on the FF testbed. The inference code, user interface code for deployment, and the demo of deployment are included in the supplementary files.

%\subsubsection{Engineering Challenges Faced in Deployment}
During the deployment of NSF-MAP in the real-world manufacturing environment at the FF lab, several challenges arise that require careful management. One of the primary concerns is adapting the code to align with the format of the input data stream, which includes both time series and image data, to ensure seamless integration and functionality. Integrating the cameras with the local PC, where the process ontology and NSF-MAP are deployed, presents an additional effort. A significant challenge is the discrepancy in the sampling rates between the OPC server and the data set used for training, necessitating efforts to synchronize these rates. Additionally, the data used for the current deployment were collected eight months prior, which may lead to slight differences in the sensor data ranges. As a result, considerable effort is devoted to adjusting and fine-tuning the code to accommodate the new data stream. The simulation of anomalies presents another challenge, as accurately replicating real-world scenarios proves difficult. Furthermore, the selection of appropriate robots and sensor values for testing, particularly in conjunction with simulated anomalies, is a task that requires close collaboration with domain specialists from the FF lab. Despite these challenges, the combined efforts and expertise of all stakeholders facilitate the successful resolution of these issues, thereby contributing to the advancement and refinement of NSF-MAP’s deployment in the manufacturing environment.

\section*{Ethical Statement}

There are no ethical issues.

\section*{Acknowledgments}
This work is supported in part by NSF grants \#2119654, “RII Track 2 FEC: Enabling Factory to Factory (F2F) Networking for Future Manufacturing” and SCRA grant “Enabling Factory to Factory (F2F) Networking for Future Manufacturing across South Carolina”.

\bibliographystyle{named}
\bibliography{ijcai25}

\end{document}